\documentclass[a4paper, 10 pt, conference]{hsmr} 
\usepackage[utf8]{inputenc}
\IEEEoverridecommandlockouts                       
\usepackage{mathtools}
\usepackage{geometry}
\usepackage[labelsep=space]{caption}
\usepackage{authblk}
\usepackage{graphicx}
\usepackage[colorlinks,urlcolor=blue]{hyperref}
\usepackage{amsfonts} 
\graphicspath{{./images/}}
\DeclareGraphicsExtensions{.pdf,.png,.jpg}

\makeatletter
\let\NAT@parse\undefined
\makeatother
\usepackage[numbers,compress]{natbib}
\bibpunct{[}{]}{,}{n}{}{;}% to get correct punctuation for bibliography
\newcommand{\argmin}{\mathop{\mathrm{argmin}}}

\title{\vspace{-20pt}
\LARGE \bf
AI-based Agents for Automated Robotic Endovascular Guidewire Manipulation
\vspace{-10pt}} 

\author{\large Young-Ho Kim$^{1}$}
\author{\large Èric Lluch$^{2}$} 
\author{\large Gulsun Mehmet$^{1}$}
\author{\large Florin C. Ghesu$^{2}$} 
\author{\large Ankur Kapoor$^{1}$ \vspace{-12pt}} 

\affil{\normalsize\textit{$^{1}$Digital Technology \& Innovation, Siemens Healthineers, Princeton, NJ, USA,} \\ 
\normalsize\textit{$^{2}$Digital Technology \& Innovation, Siemens Healthineers, Erlangen, Germany}\\ 
\small\textit{(young-ho.kim, eric.lluch, akif.gulsun, florin.ghesu, ankur.kapoor)@siemens-healthineers.com}}

\begin{document}

\maketitle
\thispagestyle{empty}
\pagestyle{empty}

\vspace*{-30pt}
\section*{INTRODUCTION}\vspace*{-5pt}

\begin{figure}[b!]
	\centering \vspace*{-15pt}
	\includegraphics[scale= 0.5]{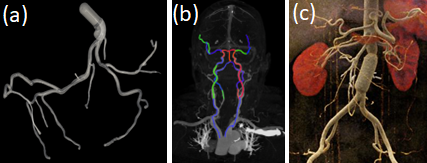}
	\vspace*{-5pt}
	\caption{3D vessel geometries are used for the clinical planning of various endovascular treatments: (a) Percutaneous Coronary Intervention (PCI) in a coronary vessel, (b) Cerebral  thrombectomy for acute ischemic stroke (AIS), (c) Transjugular intrahepatic portosystemic shunt (TIPS) for liver portal hypertension (data source of (2)(3) is \citep{goren18data})\label{fig:3d_geometry}}
\end{figure}

%1. What is endovascular guidewire manipulation?
Endovascular guidewire manipulation is essential for minimally-invasive clinical applications; Percutaneous Coronary Intervention (PCI) is used to open narrowed coronary arteries and restore arterial blood flow to heart tissue, Mechanical thrombectomy techniques for acute ischemic stroke (AIS) to remove blood clots from the brain veins, and Transjugular intrahepatic portosystemic shunt (TIPS) for liver portal hypertension use a special needle and position a wire between the portal vein through the liver. 
All procedures commonly require 3D vessel geometries from 3D CTA (Computed Tomography Angiography) images (Fig.\,\ref{fig:3d_geometry}).
During these procedures, the clinician generally places a guiding catheter in the ostium of the relevant vessel and then manipulates a wire through the catheter and across the blockage. The clinician only uses X-ray fluoroscopy intermittently to visualize and guide the catheter, guidewire, and other devices ({\em e.g.}, angioplasty balloons and stents).

%During these procedures, the clinician generally places a guiding catheter in the ostium of the relevant vessel and then manipulates a wire through the catheter and across the blockage. The clinician only uses X-ray fluoroscopy intermittently to visualize and guide the catheter, guidewire, and other devices ({\em e.g.} angioplasty balloons and stents).

Various types of endovascular robot-assisted systems\,\citep{rafii13endovascular,duan23endovascular} are being developed to provide efficient positional control of devices, helping clinicians to mitigate therapeutical risks. The primary motions that a clinician can use to control the movement and direction of the wire are rotation and pushing/retracting from the proximal end of the wire outside the insertion point on the patient's body.

Even with these robotic devices, clinicians passively control guidewires/catheters by relying on limited indirect observation ({\em i.e.}, 2D partial view of devices, and intermittent updates due to radiation limit) from X-ray fluoroscopy. Modeling and controlling the guidewire manipulation in coronary vessels remains challenging because of the complicated interaction between guidewire motions with different physical properties ({\em i.e.}, loads, coating) and vessel geometries with lumen conditions resulting in a highly non-linear system. Thus the recent literature has focused on behavior-based automated motion controls; \citet{madder17wiring} proposed the first known automatic guidewire retraction motions with a rotation of $180^{\circ}$ to cannulate a coronary artery bifurcation; In addition, robotic crossing techniques have demonstrated to generate distal guidewire motions that take advantage of the fast response of the robot system, and it's autonomous and collaborative controls\,\citep{kim2018highspeed,kim20behavior}.

%TODO: remote control: maybe also mention the advantage of being remote, eventually controlled from other centers and ultimately allowing operations for everyone (in farms, other countries, etc.)

%4. Our method: 

\begin{figure}[t!]
	\centering \vspace*{-10pt}
	\includegraphics[scale= 0.05]{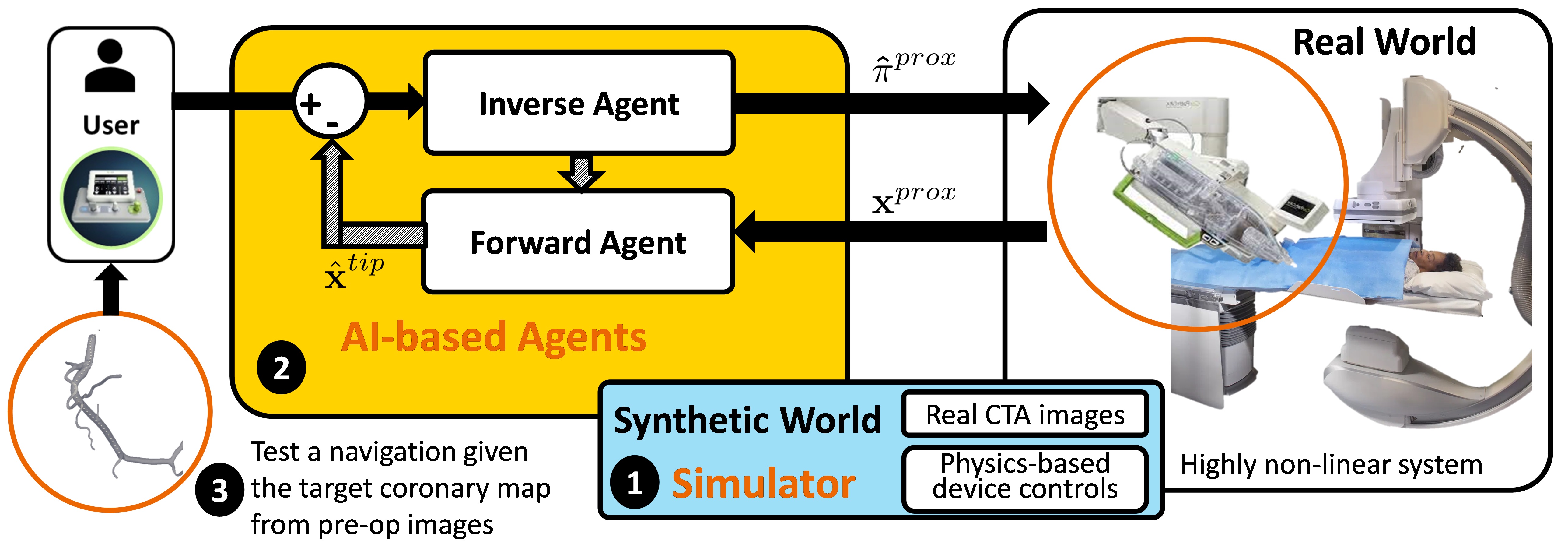}
	\vspace*{-5pt}
	\caption{An overview of the proposed AI-based agents, trained by created synthetic data using real CTA images and physics-based device controls.
  %We create scalable environment data sets from real CTA images. Each set includes segmented 3D vascular structures and centerlines with section information. Then, we apply a physics-based device control into each environment set in simulation, and create a large data set as training data. 2) We use the Transformer model to handle nonlinear sequential forward/inverse transition funtions. 3) From pre-op images, the clinician creates a desired tip state in time, then our proposed controller generates and applies the best controls for the robotic manipulation. %TODO:We should be more concise with the caption here and clearly define that the transformer model is actually modeling; what are we learning. The same in the main text where you mention the transformer.
		\label{fig:overview}}\vspace*{-20pt}
\end{figure}

This paper introduces a scalable learning pipeline to train AI-based agent models toward automated endovascular predictive device controls. Figure\,\ref{fig:overview} shows an overview of an endovascular predictive control workflow. Specifically, we create a scalable environment by pre-processing 3D CTA images, providing patient-specific 3D vessel geometry and the centerline of the coronary. Next, we apply a large quantity of randomly generated motion sequences from the proximal end to generate wire states associated with each environment using a physics-based device simulator. Then, we reformulate the control problem to a sequence-to-sequence learning problem, in which we use a Transformer-based model, trained to handle non-linear sequential forward/inverse transition functions. 

%train a Transformer model to handle non-linear sequential forward/inverse transition functions. Finally, we validate our method with a test set in simulator settings.
%We propose to reformulate the control problem to a sequence-to-sequence learning problem, which we address using a Transformer-based architecture with attention mechanisms

\section*{MATERIALS AND METHODS}\vspace*{-5pt}
%5 Proposed system overview: goal, FK IK
At the torquer's attachment point to the robot, the wire's proximal state, ${\bf x}^{prox} \equiv \langle\Delta, \phi\rangle$, has a translation $\Delta$  and rotation $\psi$. At the distal (tip) of the wire, the state ${\bf x}^{tip}$ is defined by its position and orientation in ${SE}(3)$. The physical parameters of the wire are defined as ${\bf p}$, including diameter, Poisson ratio, Young modulus, number of elements, etc. Let $ {\bf e} \in \mathcal{E}$ represent the spatial environment of the vessel, including 3D vessel geometry, 3D vessel centerline, and sectional labels.

%simplify information
Since we do not have a direct measurement for the tip of the guidewire, we simplify ${\bf x}^{tip} \in \mathbb{R}^3$ with regard to the 3D vessel centerline by projecting it into known 3D vessel centerline. Then, ${\bf x}^{tip} \approx \hat{{\bf x}}^{tip} = \langle\Delta^{tip}, \gamma^{tip}\rangle$ where $\Delta^{tip}$ and $\gamma^{tip}$ represent the translation along the centerline and the distance from the centerline to the tip, respectively.

Then our manipulation system can be treated as a forward/inverse transition function~$\mathcal{FF}$ and $\mathcal{IF}$ with control ${\bf \pi}^{prox} \in \Pi$ to transition from one state to another state given physical parameters ${\bf p}$ and the spatial environment of vessel ${\bf e}$:
{\footnotesize \vspace*{-5pt}
\begin{eqnarray} 
	{\bf \hat{x}}^{tip}_{t+1} = \mathcal{FF}({\bf \hat{x}}^{tip}_t, {\bf x}^{prox}_{t+1}, {\bf \pi}^{prox}_{t+1}; {\bf p},{\bf e}), \label{eq:FK} \\
	{\bf \hat{x}}^{prox}_{t+1}, {\bf \hat{\pi}}^{prox}_{t+1} = \mathcal{IF}({\bf \hat{x}}^{tip}_{t+1},  {\bf x}^{prox}_{t},  \pi^{prox}_{t}; {\bf p},{\bf e}) \label{eq:IK}
\end{eqnarray}
}

Given ${\bf p}$ and ${\bf e}$, manipulate the wire to the desired tip state {\footnotesize${\bf \hat{x}}^{tip}_{(t+1,\cdots,t+n)}$}, making all state transitions via ${\bf \pi}^{prox}$ and finding the sequence of proximal controls {\footnotesize${\bf \pi}^{prox}_{(t+1,\cdots,t+n)}$} that minimizes estimation cost $\mathcal{C}$. Then, the problem follows:

\vspace*{-10pt}
{\footnotesize
\begin{eqnarray}
\hat{\pi}^*=\argmin_{\pi_{(t+1,\cdots,t+n)} \in \Pi}~\mathcal{C}({\bf \pi}_t^{prox}, {\bf \hat{x}}^{tip}_{(t+1,\cdots,t+n)}|{\bf p},{\bf e}),\label{eq:problem} 
\end{eqnarray}
}
where $t$ is the current time and $n$ is the long-term time step.

%Ultimately, our problem is to find the best sequence of proximal controls, ${\bf \pi}^{prox}_{(t+1,\cdots,t+n)}$, given ${\bf p}$ and ${\bf e}$, based on the desired tip state of the wire ${\bf \hat{x}}^{tip}_{(t+1,\cdots,t+n)}$, minimizing the object cost ({\em i.e.,} errors of ${\bf \hat{x}}^{tip}$, safety boundary) over time. 

%6 Create scalable environment and Simulator explain

\begin{figure}[b!]
	\centering \vspace*{-10pt}
	\includegraphics[scale= 0.35]{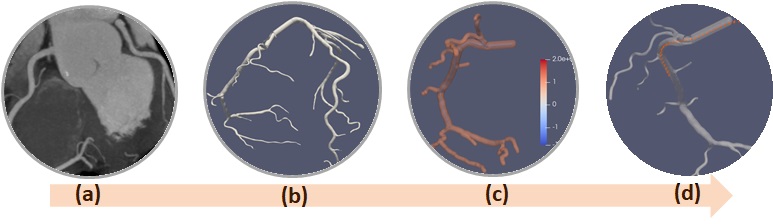}
	\vspace*{-5pt}
	\caption{Pre-processing pipeline from 3D CTA to simulated catheter insertion: (a) 3D CTA images, (b) 3D segmentation, centerline, and vessel sectional labeling, (c) Vessel boundary distance map, (d) Physics-based simulation of an inserted catheter after proximal force application \label{fig:simulation}}  
\end{figure}

Creating a scalable environment is an important step in getting a better quality of learning-based controls for intelligent agents.
Physics-based simulators are commonly used nowadays in various practical fields ({\em e.g.,} self-driving cars, pick-and-place) to improve the qualities of controls. Similarly, we create the pre-processing pipeline with 3D CTA images and 3D device simulation to create a scalable environment.
%However, due to the lack of CTA images associated with real device interaction, we create the pre-processing pipeline with real CTA images and device simulation to create the scalable environment.
%TODO The pipeline is completely automatized. I think this is importan for the story. And maybe also reporting the time of creation of 1 simulation.
Figure\,\ref{fig:simulation} shows the overall pre-processing pipeline for scalable data. First, given 3D CTA  images (Fig.\,\ref{fig:simulation}(a)), the corresponding 3D segmentation of the whole coronary vessels is computed (Fig.\,\ref{fig:simulation}(b)). Then, the 3D level set distance map that contains the minimum distance to the vessels at each 3D point is computed to define the boundary conditions (Fig.\,\ref{fig:simulation}(c)). Finally, we apply the desired force to the proximal section of the wire using the Cosserat-rod model in the simulator, which interacts with the defined boundary conditions by solving partial differential equations. %TODO describing the conservation of linear and angular momentum
Finally, this provides the scalable simulated device states inside the vessel (Fig.\,\ref{fig:simulation}(d)). 

%7. Sequential model, Transformer settings, parameters
We solve the control problem by using attention-mechanism-based learning model, Transformer\,\citep{ashish17attention}.
%We reformulate the control problem to a sequence-to-sequence learning problem, which we address using attention-mechanism-based architecture, Transformer model\,\citep{ashish17attention}. 
We use created data from the pre-processing pipeline to train the Transformer model. Our system inherently requires an open loop where spatial feedback of devices is not continuously available due to the radiation limit of X-rays. We train $\mathcal{FF}$ and $\mathcal{IF}$ as dual Transformer models to iteratively update each history of states over time. The input/output are described in Eq\,\eqref{eq:FK}\eqref{eq:IK}. Then, we can finally estimate a long sequence of state information {\footnotesize ({\em i.e.},~${\bf \hat{x}}^{tip}_{(t+1,\cdots,t+n)}$ and ${\bf \hat{x}}^{prox}_{(t+1,\cdots,t+n)}$, ${\bf \hat{\pi}}^{prox}_{(t+1,\cdots,t+n)}$)}.

%8 Experimental settings: Data collection/ Data training/ Validation
We used 100 CTA images for the right coronary artery as a simulator environment, and applied 1000 randomly generated continuous sequential controls of ${\bf \pi}^{prox}$ for 60 seconds, which generated 120 sequence sample points for each set. We used $80\%$ data as the training set and used $20\%$ for the testing set. We then present the safety ratio and difference between the estimated force and the ground-truth in the test set.

Both forward/inverse transformer models are learned with 30 sequences of states as inputs, 1 output (Many-to-one), 12 heads, 4 encoder/decoder layers, 128 dimensions of feedforward, and 0.1 dropouts. The learning rate was set to be $1e^{-3}$ for 20 epochs with Adam optimizer in Pytorch. %TODO can we justify this choice of hyperparameters?

\section*{RESULTS}\vspace*{-5pt}

\begin{figure}[t!]
	\centering \vspace*{-10pt}
	\includegraphics[scale= 0.45]{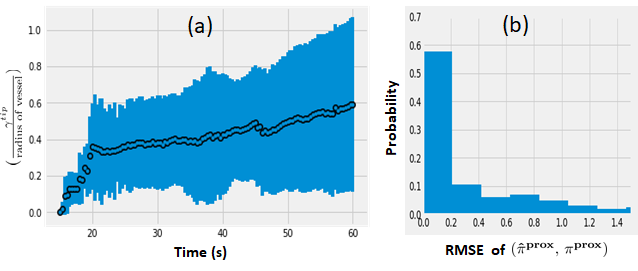}
	\vspace*{-5pt}
	\caption{Performance evaluation: (a) The safety ratio from $\frac{\gamma^{tip}}{\text{radius of vessel}}$ for operation time, and (b) Histogram of RMSE between $\hat{\pi}^{prox}$ and ${\pi}^{prox}$ \label{fig:results}}  \vspace{-20pt}
\end{figure}

We assumed the first 30 states of ${\bf \hat{x}}^{tip}$ are given as measurements. Then, we update states using $\mathcal{FF}$ and $\mathcal{IF}$ iteratively over time. %The overall tip position of root-mean-square error (RMSE) as the mean and standard deviation for $\Delta^{tip}$ shows $(0.25,0.22)~mm$ for the first 10 seconds, then later increased up to $(0.62,0.6)~mm$. 
Figure\,\ref{fig:results} (a) shows a ratio of $\frac{\gamma^{tip}}{\text{radius of vessel}}$ for the test set, where $0$ means the tip is located in the centerline while approaching to 1 represents the tip is close to the vessel wall, which means we might need to obtain new measurements from X-ray in time. The radius of the vessel that we tested is between $0.7~mm$ and $2~mm$. Figure\,\ref{fig:results} (b) shows the distribution of our control output errors by computing errors between the ground-truth of $\pi^{prox}$ and our estimation $\hat{\pi}^{prox}$.

%Then, the overall tip position of root-mean-square error(RMSE) is reported as mean and standard deviation for each $\Delta^{tip}$ and $\gamma^{tip}$. 
%First 10 seconds estimated results are $(0.25,0.22)~mm$ and $(0.09,0.1)~mm$. The next 10-30 seconds results shows $(0.31,0.27)~mm$ and $(0.12,0.14)~mm$, while the others are $(0.62,0.6)~mm$ and $(0.14,0.16)~mm$.
%Figure 4 --> errors are vessel dependent -->normalized tip error in terms of vessel diameter

\section*{DISCUSSION}\vspace*{-5pt}
Based on our results, our AI-based agents might provide an efficient approach to indicate when to turn on/off X-ray.
%leading to reduce the total radiation exposure time.
%Such intelligent agents may also be helpful for partial autonomy  in remote settings.
%Our proposed design and models are validated in a simple scenario. Also, the calibration between real and simulation settings is very important. However, our current simulator only used one known physical parameters.
As a future work, we plan to investigate more sophisticated controls for complicated scenarios with a systematic evaluation. In addition, we plan to apply various parameters to handle uncertainty.

\vspace*{-3pt}
\section*{DISCLAIMER}\vspace{-5pt}
%\vspace*{-5pt}
%\small
%This feature is based on research, and is not commercially available. Due to regulatory reasons its future availability cannot be guaranteed.
%The concepts and information presented in this paper are based on research results that are not commercially available.
{\small
The concepts and information presented in this paper are based on research results that are not commercially available. Future availability cannot be guaranteed.
}

%\nocite{*}
{
\small
\bibliography{references_hamlyn}

% Generated by IEEEtranN.bst, version: 1.13 (2008/09/30)
\begin{thebibliography}{7}
\providecommand{\natexlab}[1]{#1}
\providecommand{\url}[1]{#1}
\csname url@samestyle\endcsname
\providecommand{\newblock}{\relax}
\providecommand{\bibinfo}[2]{#2}
\providecommand{\BIBentrySTDinterwordspacing}{\spaceskip=0pt\relax}
\providecommand{\BIBentryALTinterwordstretchfactor}{4}
\providecommand{\BIBentryALTinterwordspacing}{\spaceskip=\fontdimen2\font plus
\BIBentryALTinterwordstretchfactor\fontdimen3\font minus
  \fontdimen4\font\relax}
\providecommand{\BIBforeignlanguage}[2]{{%
\expandafter\ifx\csname l@#1\endcsname\relax
\typeout{** WARNING: IEEEtranN.bst: No hyphenation pattern has been}%
\typeout{** loaded for the language `#1'. Using the pattern for}%
\typeout{** the default language instead.}%
\else
\language=\csname l@#1\endcsname
\fi
#2}}
\providecommand{\BIBdecl}{\relax}
\BIBdecl

\bibitem[Goren et~al.(2018)Goren, Avery, Dowrick, Mackle, Witkowska-Wrobel,
  Werring, and Holder]{goren18data}
N.~Goren, J.~Avery, T.~Dowrick, E.~Mackle, A.~Witkowska-Wrobel, D.~Werring, and
  D.~Holder, ``Multi-frequency electrical impedance tomography and neuroimaging
  data in stroke patients,'' \emph{Scientific Data}, vol.~5, p. 180112, 07
  2018.

\bibitem[Rafii-Tari et~al.(2013)Rafii-Tari, Payne, and
  Yang]{rafii13endovascular}
H.~Rafii-Tari, C.~Payne, and G.-Z. Yang, ``Current and emerging robot-assisted
  endovascular catheterization technologies: A review,'' \emph{Annals of
  biomedical engineering}, vol.~42, 11 2013.

\bibitem[Duan et~al.(2023)Duan, Akinyemi, Du, Ma, Chen, Wang, Omisore, Luo,
  Wang, and Wang]{duan23endovascular}
W.~Duan, T.~Akinyemi, W.~Du, J.~Ma, X.~Chen, F.~Wang, O.~Omisore, J.~Luo,
  H.~Wang, and L.~Wang, ``Technical and clinical progress on robot-assisted
  endovascular interventions: A review,'' \emph{Micromachines}, vol.~14, no.~1,
  p. 197, Jan 2023.

\bibitem[Madder et~al.(2017)Madder, Lombardi, Parikh, Kandzari, Grantham, and
  Rao]{madder17wiring}
R.~Madder, W.~Lombardi, M.~Parikh, D.~Kandzari, J.~Grantham, and S.~Rao,
  ``{Impact of a Novel Advanced Robotic Wiring Algorithm on Time to Wire a
  Coronary Artery Bifurcation in a Porcine Model},'' \emph{Journal of the
  American College of Cardiology}, vol.~70, no.~80, 2017.

\bibitem[Kim et~al.(2018{\natexlab{a}})Kim, Kapoor, Finocchi, and
  Girard]{kim2018highspeed}
Y.-H. Kim, A.~Kapoor, R.~Finocchi, and E.~Girard, ``{Evaluation of High-Speed
  Dynamic Motions for Robotic Guidewire Crossing Techniques},'' in \emph{Hamlyn
  Symposium on Medical Robotics}, London, UK, Jun. 2018.

\bibitem[Kim et~al.(2018{\natexlab{b}})Kim, Kapoor, Finocchi, and
  Girard]{kim20behavior}
------, ``An experimental validation of behavior-based motions for robotic
  coronary guidewire crossing techniques,'' in \emph{Proceedings of the
  International Symposium on Experimental Robotics}, Buenos Aires, Argentina,
  2018, pp. 14--23.

\bibitem[Vaswani et~al.(2017)Vaswani, Shazeer, Parmar, Uszkoreit, Jones, Gomez,
  Kaiser, and Polosukhin]{ashish17attention}
A.~Vaswani, N.~Shazeer, N.~Parmar, J.~Uszkoreit, L.~Jones, A.~N. Gomez, L.~u.
  Kaiser, and I.~Polosukhin, ``Attention is all you need,'' in \emph{Advances
  in Neural Information Processing Systems}, vol.~30, 2017.

\end{thebibliography}
}

\end{document}